%% file: main.tex
\documentclass[conference]{IEEEtran}
\usepackage{times}

\input{sections/preamble.tex}

\usepackage[numbers]{natbib}
\usepackage{multicol}
\usepackage[bookmarks=true]{hyperref}

\begin{document}

% paper title
\title{Network Offloading Policies for Cloud Robotics:\\ a Learning-based Approach}

% You will get a Paper-ID when submitting a pdf file to the conference system
%\author{Author Names Omitted for Anonymous Review. Paper-ID 190}

\begin{comment}
\author{\authorblockN{Michael Shell}
\authorblockA{School of Electrical and\\Computer Engineering\\
Georgia Institute of Technology\\
Atlanta, Georgia 30332--0250\\
Email: mshell@ece.gatech.edu}
\and
\authorblockN{Homer Simpson}
\authorblockA{Twentieth Century Fox\\
Springfield, USA\\
Email: homer@thesimpsons.com}
\and
\authorblockN{James Kirk\\ and Montgomery Scott}
\authorblockA{Starfleet Academy\\
San Francisco, California 96678-2391\\
Telephone: (800) 555--1212\\
Fax: (888) 555--1212}}
\end{comment}

\begin{comment}

\author{Sandeep P. Chinchali \qquad Apoorva Sharma \qquad James Harrison \qquad Amine Elhafsi \qquad Daniel Kang \\ \qquad Evgenya Pergament \qquad Eyal Cidon \qquad Sachin Katti \qquad Marco Pavone \\ Stanford University}
\end{comment}

\author{\IEEEauthorblockN{Sandeep Chinchali\IEEEauthorrefmark{1},
Apoorva Sharma\IEEEauthorrefmark{3}, 
James Harrison\IEEEauthorrefmark{4}, 
Amine Elhafsi\IEEEauthorrefmark{3}, 
Daniel Kang\IEEEauthorrefmark{1}, \\
Evgenya Pergament\IEEEauthorrefmark{2}, 
Eyal Cidon\IEEEauthorrefmark{2}, 
Sachin Katti\IEEEauthorrefmark{1}\IEEEauthorrefmark{2},
Marco Pavone\IEEEauthorrefmark{3}} 
\IEEEauthorblockA{Departments of Computer Science\IEEEauthorrefmark{1}, Electrical Engineering\IEEEauthorrefmark{2}, 
Aeronautics and Astronautics\IEEEauthorrefmark{3}, and Mechanical Engineering\IEEEauthorrefmark{4} \\
Stanford University, Stanford, CA\\
\{\texttt{csandeep, apoorva, jharrison, amine, ddkang, }\\ \texttt{evgenyap, ecidon, skatti, pavone}\}@stanford.edu}}

\maketitle

\begin{abstract}
Today's robotic systems are increasingly turning to computationally expensive models such as deep neural networks (DNNs) for tasks like localization, perception, planning, and object detection.
However, resource-constrained robots, like low-power drones,
often have insufficient on-board compute resources or power reserves to scalably run the most accurate, state-of-the art neural network  compute models.
\textit{Cloud robotics} allows mobile robots the benefit
of offloading compute to centralized servers if they are uncertain locally or want to 
run more accurate, compute-intensive models. However, cloud robotics comes with a key, often understated cost: communicating with the cloud over congested wireless networks may result in latency or loss of data. 
In fact, sending high data-rate video or LIDAR from multiple robots
over congested networks can lead to prohibitive delay for real-time applications, which we measure experimentally.
In this paper, we formulate a novel \textit{Robot Offloading Problem} ---
how and when should robots offload sensing tasks, especially if they are uncertain, to improve accuracy while minimizing the cost of cloud communication?
We formulate offloading as a sequential decision making problem for robots, and propose a solution using deep reinforcement learning. In both simulations and hardware experiments using state-of-the art vision DNNs, 
%
%offloading strategy leverages temporal coherence of input streams to achieve better prediction accuracy than all benchmark offloading strategies, while querying the cloud less than all baselines except a ``never-offload'' baseline strategy.
%
our offloading strategy improves vision task performance by between 1.3-2.6x of benchmark offloading strategies, 
allowing robots the potential to significantly transcend their on-board sensing accuracy but with limited cost of cloud communication.

%a hybrid compute strategy gets the best accuracy of the cloud but with the benefits of mostly local computation, by a considerable facto

% improves vision task performance by between
% \mpmargin{1.3-2.6x}{this might seem small, emphasize!!} of benchmark offloading solutions.
\end{abstract}

\IEEEpeerreviewmaketitle

%\input{tex/intro}

%%%%%%%%%%%%%  INTRO %%%%%%%%%%%%%%
\section{Introduction}
%%%%%%%%%%%%%%%%%%%%%%%%%%
\label{sec:intro}

\begin{figure}[t]
	\centering
   	\includegraphics[width=0.5\textwidth]{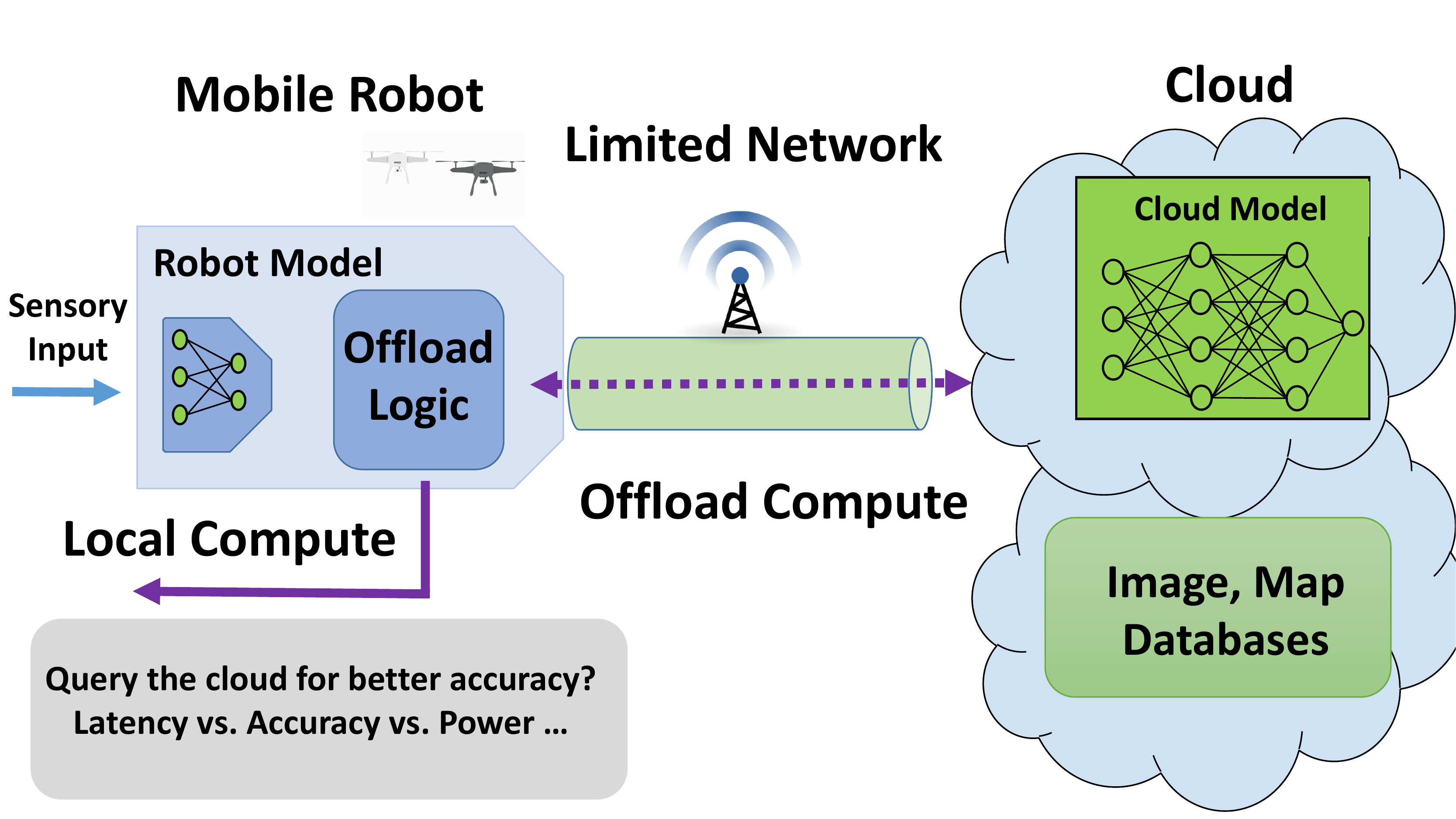}
    \caption{Autonomous mobile robots are faced with a key tradeoff. Should they rely on local compute models, which could be fast, power-efficient, but less accurate? 
    Or, should they offload computation to a more accurate model in the cloud, which increases latency due to congested networks? In this paper, we propose a novel algorithmic framework to address such tradeoffs.}
    %\caption{Autonomous mobile robots must balance local and cloud computation and judiciously use bandwidth-limited networks.}
	\label{fig:cloud}
\end{figure}

For autonomous mobile robots such as delivery drones to become ubiquitous, the amount of onboard computational resources will need to be kept relatively small to reduce energy usage and manufacturing cost. However, simultaneously, perception and decision-making systems in robotics are becoming increasingly computationally expensive\footnote{For example, deep neural network-based vision systems running on a consumer GPU are able to perform detection at a rate of approximately 180 frames per second (fps), but the GPU consumes approximately 150W. In contrast, mobile-optimized GPUs such as the Nvidia Jetson TX1 are only capable of running the detection pipeline at 70fps, while still consuming 10W of power \cite{hanDeepdrone}. As a point of comparison, consumer drones typically consume in the range of 20-150W during hover \cite{wiredDrone}. As a result, reaching practical detection rates on small mobile robots will result in large power demands that will substantially reduce the operational time of the robot.} \cite{sunderhauf2018limits}. In addition to restrictions on computation, autonomous robotic systems may also have storage limitations for, e.g., cached maps. 

To avoid these restrictions, it is possible for a robot to offload computation or storage to the cloud, where resources are effectively limitless. This approach, which is commonly referred to as \textit{cloud robotics} \cite{kuffner2010cloud}, imposes a set of trade-offs that have hitherto only been marginally addressed in the literature. Specifically, while offloading computation (for example) to the cloud reduces the onboard computing requirements, it may result in latency that could severely degrade performance, as well as information loss or total failure if a network is highly congested. Indeed, even economical querying of cloud resources may quickly overload a network in the case where the data transfer requirements are relatively large (such as high definition (HD) video or LIDAR point clouds) or where multiple robots are operating. 

In this work, we formally study the decision problem associated with offloading to cloud resources for robotic systems. Given the limitations of real-world networks, we argue that robots should offload only when necessary or highly beneficial, and should incorporate network conditions into this calculus. We view this problem as a (potentially partially-observed) Markov Decision Process (MDP) \cite{bellman1957markovian}, where an autonomous system is deciding whether to offload at every time step.

\paragraph*{Contributions and Organization} In Section \ref{sec:rw}, we survey existing work on the offloading problem in robotics and find that it under-emphasizes key costs of cloud communication such as increased latency, network congestion, and load on cloud compute resources, which in turn adversely affect a robot. We further show experimentally that current cloud robotics systems can lead to network failure and/or performance degradation, and discuss how this problem will become more severe in the future without intervention. To address this gap, we formulate a novel cost-based \textit{cloud offloading problem} in Section \ref{sec:prob}, and describe characteristics of this problem that make it difficult to solve with simple heuristics. In Section \ref{sec:approach}, we propose solutions to this problem based on deep reinforcement learning \cite{sutton1998reinforcement,szepesvari2010algorithms}, which are capable of handling diverse network conditions and flexibly trade-off robot and cloud computation. In Section \ref{sec:expts}, we demonstrate that our proposed approach allows robots to intelligently, but sparingly, query the cloud for better perception accuracy, both in simulations and hardware experiments. %Finally, we discuss extensions to this work to bridge the robotics and the network systems research communities. 
To our knowledge, this is the first work that formulates the general cloud offloading problem as a sequential decision-making problem under uncertainty and presents general-purpose, extensible models for the costs of robot/cloud compute and network communication.

% james version
%We describe existing work on this problem in both the robotics and networks research communities in Section \ref{sec:rw}. In Section \ref{sec:prob}, we formally describe the \textit{cloud offloading problem}, and describe characteristics of this problem that make it difficult to solve with simple heuristics. In Section \ref{sec:approach},  we propose solutions to this problem based on reinforcement learning, which are capable of handling diverse network conditions and robotic systems. We demonstrate our proposed approach in simulations and experimentally in Section \ref{sec:expts}. Finally, we discuss extensions to this work to bridge the robotics and the network systems research communities. 

\section{Background \& Related Work}
%%%%%%%%%%%%%%%%%%%%%%%%%%
\label{sec:rw}
\subsection{Cloud Robotics}
Cloud robotics has been proposed as a solution to limited onboard computation in mobile robotic systems, and the term broadly refers to the process of offloading to cloud-based computational resources \cite{kuffner2010cloud, goldberg2013cloud, cloudStatus, kehoe2015survey}. 
For example, a robot may offload video processing and associated perception, audio and natural language processing, or other sensory inputs such as LIDAR. Concretely, this approach has been used in mapping \cite{mohanarajah2015cloud} and localization \cite{riazuelo2014c2tam}, perception \cite{salmeron2015tradeoff}, grasping \cite{kehoe2013cloud}, visuomotor control \cite{wu2013cloud}, speech processing \cite{sugiura2015rospeex}, and other applications \cite{rahman2016cloud}. For a review of work in the field, we refer the reader to \cite{kehoe2015survey}.
Cloud robotics can also include offloading complex decision making to a human, an approach that has been used in path planning \cite{higuera2012socially, jain2015planit}, and as a backup option for self-driving cars in case of planning failures \cite{wiredStarsky}.

In general, the paradigm is useful in any scenario in which there is a tradeoff between performance and computational resources. A notable example of this tradeoff is in perception, a scenario we use as a case-study in this paper. Specifically, vision Deep Neural Networks (DNNs) are becoming the de facto standard for object detection, classification, and localization for robotics. However, as shown in Table \ref{fig:DNN_table}, different DNNs offer varied compute/accuracy tradeoffs. Mobile-optimized vision DNNs, such as MobileNets \cite{mobilenetcpu} and ShuffleNets~\cite{xiangyu2017shufflenet}, often sacrifice accuracy to be faster and use less power. While MobileNet has lower accuracy, it has significantly fewer parameters and operations than the more accurate Mask R-CNN model, and thus might be favored for an on-robot processing model. A cloud-robotics framework would give improved performance by allowing the robot to query a cloud server running the compute-intensive Mask R-CNN model as needed, and using the onboard model when the lower accuracy is tolerable. 

\begin{comment}
An example of a robot perception model is a vision deep neural network (DNN), which can localize and identify objects in static images and video. The use of vision DNNs, either on a robot or the cloud, comes with important design tradeoffs. As exemplified in Table \ref{fig:DNN_table}, the most accurate models have more parameters and operations, which increase power consumption and lead to slow inference times. However, mobile-optimized vision DNNs, such as MobileNets \cite{mobilenetcpu} and ShuffleNets~\cite{xiangyu2017shufflenet}, often sacrifice accuracy to be faster and use less power. Finally, running vision DNNs on a CPU often has per-image inference times above 200 milliseconds \cite{mobilenetcpu}, while GPUs on optimized embedded platforms can run the same models in less than 50 ms \cite{TFTRT} but are typically more costly.
In essence, a roboticist must trade off the accuracy needs of a sensing task with limits on compute cost, inference latency, and power usage, which could, for example, limit the effective range of delivery drones. 
A key contribution of our work is to distill the above tradeoffs into a cost-based cloud offloading decision framework. 
\end{comment}

\subsection{Costs of Offloading}
While offloading computation or storage to the cloud has the potential to enable cheap mobile robots to perform increasingly complex tasks, these benefits come at a cost. Querying the cloud is not instant, and there are costs associated with this latency. Furthermore, mobile robots largely use wireless networks (e.g., cellular or WiFi networks), which can be highly stochastic and low
bandwidth~\cite{commutewifi}. Often, the offloaded data can be large relative to this bandwidth: HD video
(e.g., for detecting obstacles) \textit{from a single robot} can be over 8 megabits per second
(Mbps)~\cite{bitrate}, while cellular networks are often uplink-limited and have
between 1-10 Mbps~\cite{commutewifi,mao2017neural} to share \textit{across} users.

\begin{table}[t]
    \centering
    \vspace{2mm}
	\footnotesize {
	\begin{tabular}{ |p{1.8cm}|>{\columncolor{TealBlue!20}}c|>{\columncolor{YellowGreen!20}}c|>{\columncolor{CornflowerBlue!20}}c| c| }
		\hline
		\textbf{DNN} & \textbf{Accuracy} & \textbf{Size} & \textbf{CPU Infer.} & \textbf{GPU Infer.}\\ 
		%\hline
		MobileNet v1  & 18 & 18 MB & 270 ms & 26 ms \\
		\hline
		MobileNet v2  & 22 & 67 MB & 200 ms & 29 ms \\
		\hline
		\maskrcnn & 45.2 & 1.6GB & 325 ms & 18 ms \\
		\hline
	\end{tabular}
	}
	\caption{Accuracy, size, and speed tradeoffs of deep neural networks, where accuracy is the standard mean average precision (mAP) metric on the MS-COCO visual dataset \cite{coco}.}
	\label{fig:DNN_table}
\end{table}

Current state-of-the-art methods in cloud robotics largely fail to adequately consider these costs. For example, to limit network bandwidth utilization, \cite{mohanarajah2015cloud} offload only key-frames (as opposed to all data) in mapping and localization. These key-frames are determined without considering the state of the network connection, and are sent at a predetermined frequency. In \cite{rahman2016cloud}, the authors factor in the current state of the system, and hand-design a one-stage decision rule. However, designing such decision rules involves a number of trade-offs and can be difficult. In \cite{salmeron2015tradeoff}, the authors present a detailed comparison of offloading characteristics in cloud robotics to inform this design process, however, hand engineering heuristic solutions remains difficult and very domain specific, and it is unclear if these approaches can scale to higher data requirements where the costs of offloading become more significant.

Related offloading architectures have been employed outside of cloud robotics in the Internet-of-Things (IoT) community, especially as machine learning models have increased in complexity \cite{pakha, hotnets}. However, the offloading techniques used in this field rely on similar techniques, e.g., determining and offloading key-frames for object detection and utilizing heuristics such as frame differences to select the key-frames \cite{chen2015glimpse}; the policy is not optimized with system-level performance in mind. Alternative approaches include splitting neural network models across edge devices (in our application, the robot) and the cloud \cite{kang2017neurosurgeon}, but these may perform poorly under variable network conditions that are likely as a mobile robot navigates through regions of varying signal quality. 

In this paper, we present an approach to address these hitherto under-emphasized costs in cloud robotics, that incorporates features of the input stream and network conditions in the system-level decision problem of whether or not to offload.

\begin{figure}[t]
    \centering
    \subfloat{
        \includegraphics[width=0.5\textwidth]{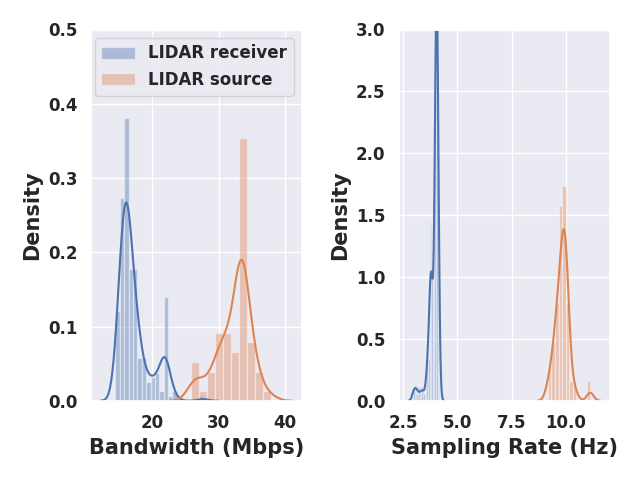}
    }
    \caption{Streaming LIDAR over WiFi using the Robot Operating System (ROS) produces high-bitrate point cloud streams, which can lead to a sender receiving only half the data (as pictured) or even network failure in the multi-robot setting.}
    \label{fig:ROS_lidar}
\end{figure}

\subsection{Case Study: Costs of Offloading LIDAR Data}

% SANDEEP
\paragraph*{Network experiments with ROS}
To motivate our contributions, we measure the impact of streaming LIDAR over a real network, which might occur in a cloud-based mapping scenario. The hurdles of streaming LIDAR over wireless networks have previously been informally described in online forums \cite{ROSate}, but, to our knowledge, never rigorously measured in academic literature. We measured the effective data-rate when sending LIDAR over a WiFi connection between an embedded compute platform useful for robotics (the \nvidia{} \jetson{}) and a central server.
A Velodyne VLP-16 LIDAR was connected to the source (\nvidia{} \jetson{}) and produced point clouds of average size $426$ Kilobytes (KB) at a rate of 10 Hz, as expected from Velodyne specifications. Using the standard Robot Operating System (ROS) \cite{ROS} as the message passing interface, point clouds were sent in real-time on an \textit{uncongested} wireless network, with only one other dormant connected machine, to a central server.

At the source, we measured a median $33.3$ Mbps data-rate, as expected from our measured average data size of $426$ (KB) KiloBytes at 10 Hz. However, at the receiver, we measured a median data-rate of $16.4$ Mbps (less than half of the sender's) with a received sampling frequency of only $3.92$ Hz. Fig. \ref{fig:ROS_lidar} contrasts sender (red) and receiver (blue) data-rates.
% data size is large, similar to that for 4k video, matches point cloud measurements from other sources
% hard for one network, theoretically can handle, but often between 1-10 Mbps based on congestion, will not scale to several robots
% indeed need Gigabit ethernet, which would effectively ground the robots
Since the WiFi link had regular delay statistics, we attribute the low received data-rate to inefficiencies of ROS in handling a large ($33.3$ Mbps) \textit{sustained} stream. In fact, official ROS documentation for the bandwidth measurement tool \texttt{rostopic bw} acknowledges that poor network connectivity and Python, not faster C++, code could be the cause of the receiver not keeping pace with the sender.
Though anecdotal, we noticed several users on ROS forums with similar issues for both LIDAR and video.\footnote{The post \textit{ROS Ate My Network Bandwidth!} details similar \cite{ROSate} behaviors.}
%Further, our measurements of point cloud size resemble those of other users \cite{ROSate,pcloudsize}. 

For a single sender-receiver pair, the problem we measured in Fig. \ref{fig:ROS_lidar} may be solved by optimizing ROS receiver code. 
%For a single sender-receiver pair, the problem of dropped LIDAR packets at the sender may be solved by optimizing ROS receiver code. 
Or, one could state-fully encode differences in LIDAR point clouds, inspired by today's video encoders \cite{videoencoding}.
However, the point cloud stream of $33.3$ Mbps is disconcertingly large, since WiFi networks often attain only 5-100 Mbps \cite{verma2013wifi,lifewirewifi} while uplink-limited cellular networks often only sustain 1-10 Mbps \cite{verizon,commutewifi,mao2017neural} across users due to congestion or fading environments.

Indeed, to stress test the above scenario, we streamed data from several Velodyne sensors over a previously uncongested WiFi network and observed severe delay, dropped ROS messages, and network outages before we could take rigorous measurements.
%However,

%Our experiments are just a first glimpse of the large network impacts of cloud robotics, which mirror problems in streaming HD video to the cloud observed in the computer systems community 
Our experiments have striking resemblance to issues faced in the computer systems community with streaming HD video to the cloud for computer vision \cite{pakha, hotnets, droneMPC}.
%As mobile robot swarms venture into areas without stable WiFi links and instead offload over cellular, this will only exacerbate problems due to lower bandwidth and higher latency \cite{commutewifi}.
In the context of robotics, an informal report from Intel estimates that self-driving cars will generate 4 Terabytes of sensor data per day, much more than served by today's cell networks \cite{intel}. Even if this data could be streamed to the cloud, it will place an enormous load on cloud compute services, such as the recent widespread outage of Amazon's Alexa speech-processing agent due to an influx of new devices on Christmas day \cite{alexadebacle}.
As more robotics platforms turn to the cloud, such as the Anki toy robot which offloads the bulk of its interactive speech processing \cite{ankiforbes}, robots will have a huge incentive to minimize their network impact. 

Indeed, sharing the network will allow swarms of robots to reap the benefits of the cloud and, importantly, reduce the latency they experience in receiving informative responses from the cloud due to network congestion. We now propose an algorithmic framework on how robots can combine local compute, active data filtering, and beneficial querying of the cloud. 
%\jhtodo{add reference to alexa?}

%\section{Motivation for L}
\section{Problem Statement}
%%%%%%%%%%%%%%%%%%%%%%%%%%%
\label{sec:prob}

In this paper, we focus on an abstract cloud robotics scenario, in which a robot experiences a stream of sensory inputs that it must process. At each timestep, it must choose to process the input onboard or to offload the processing to the cloud over a network. In this section, we offer practically motivated abstractions of these ideas, and combine them to formally define the robot-offloading problem.

\begin{figure}[t]
    \centering
	\subfloat{
    	\includegraphics[width=0.5\textwidth]{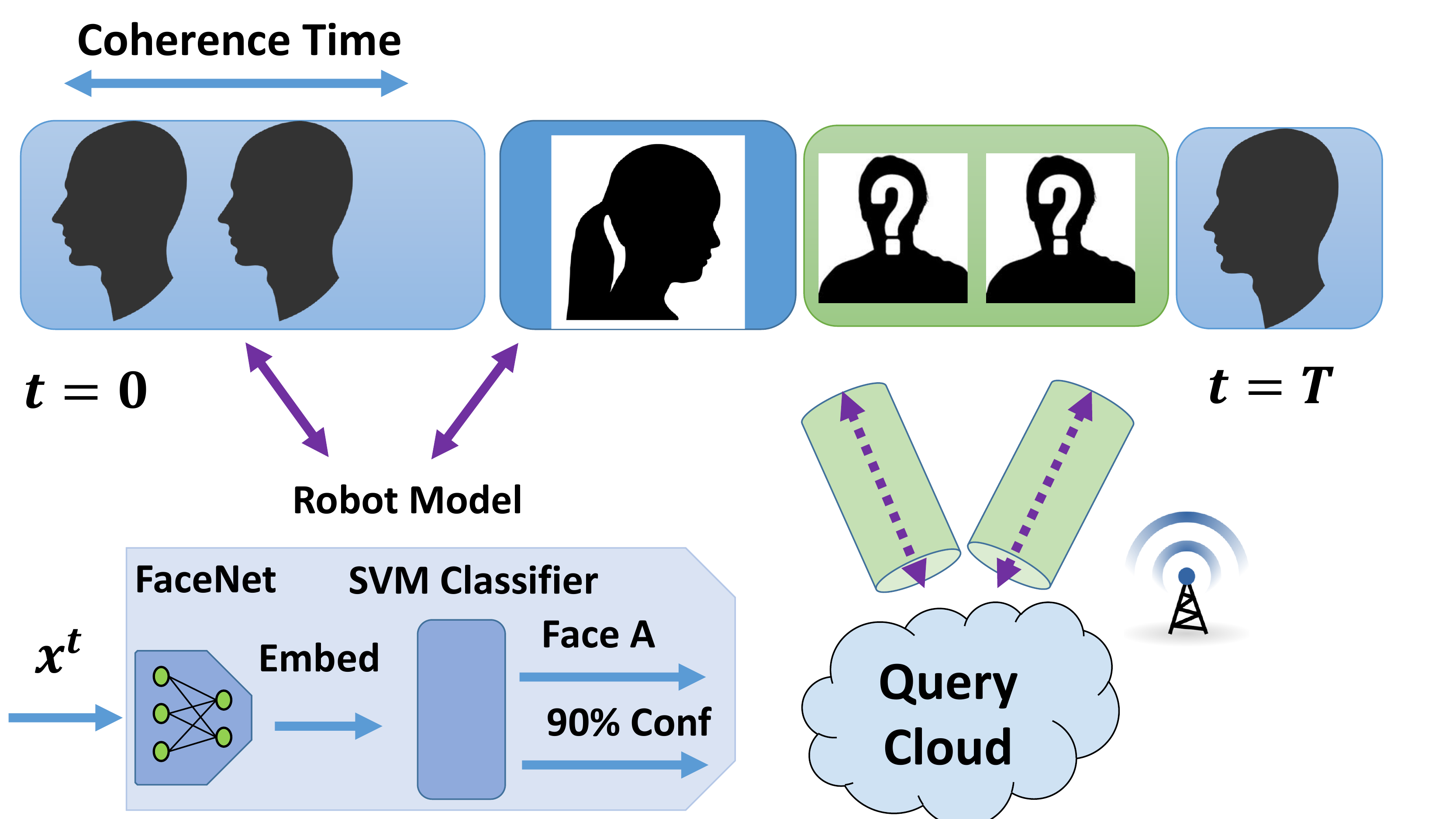}
    }
    %\caption{An episode of the streaming face recognition scenario.}
    \caption{While our framework is general, we demonstrate it on face recognition from video, a common task for personal assistance robots or search-and-rescue drones. Video surveillance occurs over a finite horizon episode where a robot can use either an optimized local model or query the cloud if uncertain.}
    \label{fig:facenet_demo}
\end{figure}

\paragraph*{Sensory Input}
We model the raw sensory input into the robot as the sequence $\{x^t\}_{t=0}^T$, where $x_t$ represents the data, such as a video frame or LIDAR point cloud, that arrives at time $t$. 
While the robot cannot know this sequence in advance, there may be properties of the distribution over these inputs that may guide the robot's offloading policy. For example, this stream may have temporal coherence (see Fig. \ref{fig:facenet_demo}), such as when objects are relatively stationary in video \cite{noscope,videoencoding}, which implies that $x^{t}$ is similar to $x^{t-1}$. As an example, a person will rarely appear in a video for only a single frame, and instead will be present for an extended time period, and the image of the person will likely change slowly. However, building a model of coherence can be difficult, and so a model-based approach even based on relatively simple heuristics may be difficult. Instead, we turn to model-free approaches, which sidestep modelling temporal coherence and instead can directly learn (comparatively) simple decision rules. 

\paragraph*{Computation Models}
The computation that we consider offloading to the cloud is the process of estimating some output $y^t$ given some input $x^t$. For example, in the scenario of processing a video stream, $y^t$ could be a semantically separated version of the input frame $x^t$ (e.g., object detection), useful for downstream decision making. For the sake of generality, we make no assumptions on what this computation is, and only assume that both the robot and the cloud have models that map a query $x^t$ to predictions of $y^t$ and importantly, a score of their confidence $\yconf^t$:
\begin{align*}
    \yhatrobot^t, \yconfrobot^t &= \frobot(x^t) \\
    \yhatcloud^t, \yconfcloud^t &= \fcloud(x^t).
\end{align*}
Typically, $\frobot$ is a computationally efficient model suitable for resource-constrained mobile robots. In contrast, $\fcloud$ represents a more accurate model which cannot be deployed at scale, for example a large DNN or the decision making of a human operator. The accuracy of these models can be measured through a loss function $\loss$ that penalizes differences between the predictions and the true results, e.g., the cross entropy loss for classification problems or root mean squared error (RMSE) loss for regression tasks. In the experiments in this paper, we operate in a classification setting, in which confidences are easy to characterize (typically via softmax output layers). However in the regression setting, there are also a wide variety of models capable of outputting prediction confidence \cite{blundell2015weight,gal2017deep,harrison2018meta}. The use of separate, modular robot and cloud models allows a robot to operate independently in case of network failure.

\paragraph*{Offload Bandwidth Constraints}
The volume of data that can be offloaded is limited by bandwidth, either of the network, or a human operator. We abstract this notion by giving the robot a finite query budget of $\budget$ samples $x^t$ that a robot can offload over a finite horizon of $T$ timesteps. This formalism flexibly allows modeling network bandwidth constraints, or rate-limiting queries to a human. Indeed, the fraction of samples a robot can offload in finite horizon $T$ can be interpreted as the robot's ``fair-share'' of a network link to limit congestion, a metric used in network resource allocation \cite{TCP,padhye}.

These factors impose a variety of tradeoffs to consider when designing an effective offloading policy. Indeed, we can see that the problem of robot offloading can be seen as a sequential decision making problem under uncertainty. Thus, we formulate this problem as a Markov Decision Process (MDP), allowing us to naturally express desiderata for an offloading policy through the design of a cost function, and from there guide the offloading policy design process.

\subsection{The Robot Offloading Markov Decision Process}

\begin{figure}[t]
	\centering
	\includegraphics[width=\columnwidth]{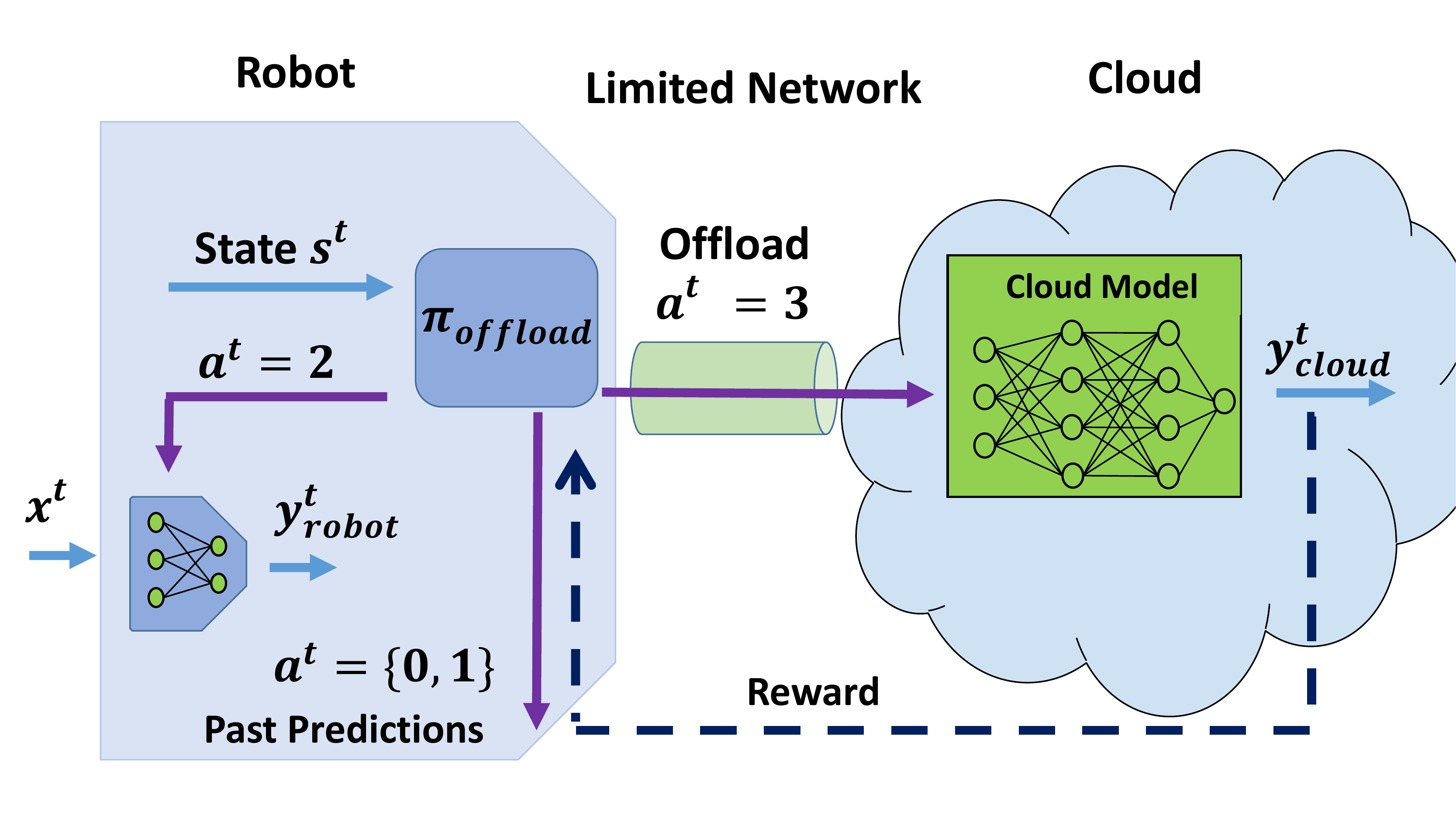}
    \caption{We formulate a novel Robot Offloading MDP, depicted above, where a robot
    uses an on-board offloading policy to select if it should use cached predictions, query a local model, or
    incur the cost, but also accuracy benefits, of querying the cloud.}
	\label{fig:offload_MDP}
\end{figure}

In this section, we express the generic robot offloading problem as an MDP 
\begin{equation}
\label{eq:offloading_mdp}
\MDPoffload = (\Soffload, \Aoffload, \Roffload, \Poffload, T),
\end{equation}
where $\Soffload$ is the state space, $\Aoffload$ is the action space, $\Roffload : \Soffload \times \Aoffload \rightarrow \mathbb{R}$ is a reward function, $\Poffload : \Soffload \times \Aoffload \times \Soffload \rightarrow [0,1]$ defines the stochastic dynamics, and $T$ is the problem horizon. In the following section, we define each of these elements in terms of the abstractions of the robot offloading problem discussed earlier. Figure \ref{fig:offload_MDP} shows the interplay between the agent (the robot), the offloading policy, and the environment, consisting of the sensory input stream and the robot and cloud prediction models. 

% ACTION
%%%%%%%%%%%%%%%%
\paragraph*{Action Space}
We consider the offloading decision problem to be the choice of which prediction $\yhat$ to use for downstream tasks at time $t$. The offloading system can either (A) choose to use past predictions and exploit temporal coherence to avoid performing computation on the new input $x^t$, or (B) incur the computation or network cost of using either the on-device model $\frobot$ or querying the cloud model $\fcloud$. Specifically, we have four discrete actions:

\begin{align}
\small
    \label{eq:twostage_action}
    \aoffload^t = 
        \begin{cases}
        0, ~\mathrm{use~past~robot~prediction}~ \yhat^t = \frobot(x^{\taurobot})  \\
        1, ~\mathrm{use~past~cloud~prediction}~ \yhat^t = \fcloud(x^{\taucloud}) \\
        2, ~\mathrm{use~current~robot~prediction}~ \yhat^t = \frobot(x^{t}) \\
        3, ~\mathrm{use~current~cloud~prediction}~ \yhat^t = \fcloud(x^{t}) \\
        \end{cases}
\end{align}
where $\taurobot < t$ is the last time the robot model was queried, and $\taucloud < t$ is the last time the cloud model was queried.

% STATE
%%%%%%%%%%%%%%%%%%
\paragraph*{State Space}
We define the state in the offload MDP to contain the information needed to choose between the actions outlined above. 
Intuitively, this choice should depend on the current sensory input $x^t$, the stored previous predictions, a measure of the ``staleness'' of these predictions, and finally, the remaining query budget. We choose to measure the staleness of the past predictions by their age, defining $\Delta \trobot = t - \taurobot$ and 
$\Delta \tcloud = t - \taucloud$.
%\begin{align*}
%    \Delta \trobot &= t - \taurobot \\
%    \Delta \tcloud &= t - \taucloud 
%\end{align*}
Formally, we define the state in the offloading MDP to be:

\begin{align}
\small
    \label{eq:twostage_state}
    \soffload^t = [\underbrace{\phi(x^t)}_{\mathrm{features~of~input}}, \underbrace{\frobot(x^{\taurobot})}_{\mathrm{past~robot}}, \underbrace{\fcloud(x^{\taucloud})}_{\mathrm{past~cloud}}, \\ \nonumber 
    \underbrace{\Delta \trobot}_\mathrm{last~robot~query}, \underbrace{\Delta \tcloud}_{\mathrm{last~cloud~query}}, \underbrace{\Delta \budget}_\mathrm{remaining~queries}, \underbrace{T-t}_{\mathrm{time~left}}].
\end{align}

Note that the sensory input $x^t$ may be high-dimensional, and including it directly in the planning problem state could yield an extremely large state-space. Instead, we consider including features $\phi(x^t)$ that are a function of the inputs. We note that in place of our choice of input representation, these state elements may be any summary of the input stream. The specific choice is context dependent and depends on the expense associated with utilizing the chosen features, as well as standard encodings or feature mappings. We describe the choice of practical features $\phi$ in Section \ref{sec:expts}.

% % ACTION
% %%%%%%%%%%%%%%%%%%
% \noindent \textbf{Action $\aoffload^t \in \Aoffload = \{0,1,2,3\}$:}\\
% Four offloading actions represent whether to (A) 
% exploit the temporal coherence of inputs and simply use past predictions to avoid
% compute cost or (B) incur the cost of querying edge or cloud perception models for the new input $x^t$.
% We denote the prediction \textit{output by the offloading system}, which could come from the robot or cloud model depending on which prediction was used,  as $\yhat^{t}$. 

% Importantly, if the offloader chooses $\aoffload^t = 3$ when it has exceeded its cloud query-budget, i.e $\Delta \budget = 0$,
% the executed action is to query the edge model ($\aoffload^t = 2$), which represents an actuation limit. A hard
% constraint barring excess cloud queries is easy to implement in offloader control software.

% More complex actions are associated with a higher cost, denoted by $\cost(\aoffload^t) \in \reals$, where
% \begin{align*}
% \cost(\aoffload^t = 3) > \cost(2) > \cost(1) \ge \cost(0) \ge 0.
% \end{align*}
% Actions that use past predictions without querying any model have equal cost, so $\cost(1) = \cost(0)$.
% In practice, costs can include inference power and latency, guided by metrics such as those
% in Figure \ref{fig:DNN_table}.

% DYNAMICS
%%%%%%%%%%%%%%%%%%
\paragraph*{Dynamics}
The dynamics in the robot offloading MDP capture both the stochastic evolution of the sensory input, as well as how the offloading decisions impact the other state elements such as the stored predictions and the query budget. The evolution of $x^t$ is independent of the offloading action, and follows a stochastic transition model that is domain-specific. For example, the evolution of video frames or LIDAR point clouds depends on the coherence of the background scene and robot mobility. The time remaining $T-t$ deterministically decrements by 1 at every timestep. The other state variable's transitions depend on the chosen action. 

If $\aoffload^t \in \{0,1\}$, then the past prediction elements of the state do not change, but we increment their age by one. If $\aoffload^t = 2$, meaning we used the current on-robot model, then we update the stored robot model prediction $\frobot$ and reset its age to $\Delta \trobot = 0$. Similarly, if we choose to query the cloud model, $\aoffload^t = 3$, then we update the stored $\fcloud$ prediction and reset its age to $\Delta \tcloud = 0$, and also decrement the query budget $\budget$ by 1.

The modelling of the network query budget is directly based on by our measurements (Fig. \ref{fig:ROS_lidar}) and recent work in the systems community on network congestion \cite{commutewifi, pakha, hotnets}. Our use of sequential features is inspired by the coherence of video frames \cite{noscope, videoencoding}, which we also measured experimentally and observed for LIDAR point clouds. 

% If $\aoffload^t \in \{0, 1, 2\}$, the last prediction output $\yhat^{t-1}$ is updated based on whether a past robot, past cloud, or current robot prediction was used. If $\aoffload^t = 3$, meaning the sample was offloaded to the cloud, $\yhat^{t-1}$ is updated \textit{and} $\Delta \budget$ is decremented by one indicating we have used a scarce cloud query. 

% In all cases, input $x^t$ and features $\phi(x^t)$ evolve stochastically based on the input stream and time remaining $\Delta t = T-t$ decrements by one each timestep.
% Further, we update past predictions $\frobot(x^{\taurobot^L})$ and $\fcloud(x^{\taucloud^K})$ in the state $\soffload^t$ 
% whenever a robot or cloud model is queried using $\aoffload^t \in \{2,3\}$, as well as the time gaps $\Delta \trobot$ and $\Delta \tcloud$.

% REWARD
%%%%%%%%%%%%%%%%%%
\paragraph*{Reward}
We choose the reward function in the MDP to express our objective of achieving good prediction accuracy while minimizing both on-robot computation and network utilization. We can naturally express the goal of high prediction accuracy by adding a penalty proportional to the loss function $\loss$ under which the cloud and robot models are evaluated. We note, however, that this choice of loss is arbitrary, and a loss derived from some downstream application may be used instead. Indeed, if a scenario is such that mis-classification will result in high cost (e.g., mis-classifying a human as a stationary object during path planning), this may be incorporated into the MDP reward function. To model the cost of network utilization and computation, we add action costs. This gives us the reward function
    % Reward $\Roffload$ trades off a penalty for prediction accuracy with the tiered cost for querying models described above.
    % During training, we assume we have the ground truth label $y^t = \yoracle^t$ so we can compute a loss function $-\mathcal{L}(\yoracle^t, \yhat^t)$ relative to prediction $\yhat^t$, which is used in the reward component for prediction accuracy. 
\begin{align}
\small
    \label{eq:twostage_reward}
    \Roffload^t(s^t,a^t) = 
          -\alphaaccuracy \underbrace{\loss}_{\mathrm{model~error}} - \underbrace{\betacost~\mathtt{cost}(a^t)}_{\mathrm{latency,~compute}},
\end{align}
where $\alphaaccuracy$, $\betacost$ are weights. The costs for network utilization are best derived from the economic analysis of onboard power usage and the cost of bandwidth utilization. For example, a mobile robot with a small battery might warrant a higher cost for querying the onboard model than a robot with a higher battery capacity.

\subsection{The Robot Offloading Problem}
Having formally defined the robot offloading scenario as an MDP, we can quantify the performance of an offloading policy in terms of the expected total reward it obtains in this MDP. This allows us to formally describe the general robot offloading problem as:

\begin{problem}[Robot Offloading Problem]
\label{problem:mobile_offloading_problem}
Given robot model $\frobot$, cloud model $\fcloud$, a cloud query budget of $\budget$ over a finite horizon of $T$ steps, and an offloading MDP $\MDPoffload$ (Equation \ref{eq:offloading_mdp}), 
find optimal offloading control policy $\pioffload^{*}: \Soffload \rightarrow \Aoffload$ that maximizes expected cumulative reward $\Roffload$:

\begin{align}
\label{eq:mobile_offloading_problem}
\pioffload^*\in \argmax_{\pioffload} \expec_{x^0, \ldots, x^T}\left(\sum_{t=0}^{T} \Roffload(\soffload^t, \aoffload^t)\right),
\end{align}
where $\aoffload^t = \pioffload^{*}(\soffload^t)$. 
\end{problem}

% We now explore several solutions to Problem \ref{problem:mobile_offloading_problem} and simulate them using real computer vision DNNs.

% \subsection{Practical MDP Formulation}
Our MDP formulation, depicted in Fig. \ref{fig:offload_MDP}, is based both on experimental insights and practical engineering abstractions. A key abstraction is the use of separate, modular robot and cloud perception models. 
Thus, a designer can flexibly trade-off accuracy, speed, and hardware cost, using a suite of pre-trained models available today \cite{mobilenetcpu}, as alluded to in Table \ref{fig:DNN_table}. Importantly, the robot can always default to its local model in case of network failure, which provides a guarantee on minimum performance.

While we have framed this problem as an MDP, we cannot easily apply conventional tools for exactly solving MDPs such as dynamic programming, as many of the aspects of this problem are hard to analytically characterize, notably the dynamics of the sensory input stream. This motivates studying approximate solution techniques to this problem, which we discuss in the following section.

We emphasize that the framework we present is agnostic to the sensory input modality, and is capable of handling a wide variety of data streams or cost functions. Moreover, the action space can be simply extended if multiple offloading options exist. As such, it describes the generic offloading problem for robotic systems.

\section{A Deep RL Approach to Robot Offloading}
%%%%%%%%%%%%%%%%%%%%%%%%%%
\label{sec:approach}
%\subsection{Our Approach}
\paragraph*{Our Approach}
%Our approach to the offloading problem is based on deep reinforcement learning %(RL) \cite{sutton1998reinforcement, szepesvari2010algorithms, mnih2013playing}. %We refer to the policy trained via RL as $\pioffloadRL$. 
%We turn to reinforcement learning for several reasons. 
We approach the offloading problem using deep reinforcement learning (RL)
\cite{sutton1998reinforcement, szepesvari2010algorithms, mnih2013playing}
for several reasons. 
First and foremost, model-free policy search methods such as reinforcement learning avoid needing to model the dynamics of the system. While most of the dynamics of the system are relatively simple, it is extremely difficult to model the evolution of the incoming sensory inputs. The model-free approach is capable of learning optimal offloading policies based solely on the features included in the state, and may avoid trying to predict incoming images, for example. Moreover, the use of a recurrent policy allows better estimation of latent variables defining the context of the incoming images. This recurrent policy accounts for possible non-Markovianity of the state. Indeed, since the state vector only includes features from the previous two most recent inputs, a Markovian policy likely can not accurately model the sensory stream.

There are several other advantages to using RL to compute good offloading policies. RL enables simple methods to handle stochastic rewards. We have chosen a relatively general reward function in the previous section, which may be stochastic due to variable costs associated with network conditions or variable cost of computation due to other processes. Finally, an RL based approach allows inexpensive evaluation of the policy, as it is not necessary to evaluate dynamics and perform optimization-based action selection as in, e.g., model predictive control \cite{Camacho2013}. In contrast to these approaches, a deep RL-based approach requires only evaluating a neural network. Because this policy evaluation is performed as an intermediate step to perception onboard the robot, efficient evaluation is critical to achieving low latency. 

We represent the RL offloading policy as a deep neural network and train it using the 
Advantage Actor-Critic (A2C) algorithm \cite{mnih2016asynchronous}. We discuss the details of the training procedure in the next section. We refer to the policy trained via RL as $\pioffloadRL$.
%There, we also discuss concise feature extractors of the state, training specifications, and our resulting lightweight RL policy network which is much smaller than even mobile vision DNNs.  

\paragraph*{Baseline Approaches}

We compare the RL-based policy against the following baseline policies: 

\begin{enumerate}
    \item \textbf{Random Sampling $\pioffloadrandom$}

    This extremely simple benchmark chooses a random $\aoffload^t \in \{0,1,2,3\}$ when the cloud query budget
    is not saturated and, afterwards, chooses randomly from actions $0-2$.

    \item \textbf{Robot-only Policy $\pioffloadrobot$}

    The robot-only policy chooses $\aoffload^t = 2$ at every time-step to query the robot model and can optionally use past robot predictions $\aoffload^t = 0$ in between. 
    \item \textbf{Cloud-only Policy $\pioffloadcloud$}
    The cloud-only policy chooses $\aoffload^t = 3$ uniformly every $\frac{\budget}{T}$ steps (queries the cloud model) and uses the past cloud predictions $\aoffload^t = 1$ in between. Essentially, we periodically sample the cloud model and hold the prediction.

    \item \textbf{Robot-uncertainty Based Sampling $\pioffloadheuristic$}
    This policy uses robot confidence $\yconfrobot^t$ to offload the $q^{th}$ percentile least-confident samples
    to the cloud as long as the remaining cloud query budget allows.

\end{enumerate}

While approaches 2 and 3 may seem simple, we note that these are the de-facto strategies used in either standard robotics (all robot computations) or standard cloud robotics (all offloading with holds to reduce bandwidth requirements). Robot-uncertainty based sampling is a heuristic that may be used for key-frame selection, analogously to \cite{mohanarajah2015cloud}. 

\section{Experimental Performance of \\Our Deep RL Offloader}
%%%%%%%%%%%%%%%%%%%%%%%%%%
\label{sec:expts}
\input{sections/evaluation.tex}

\section{Discussion and Conclusions}
%%%%%%%%%%%%%%%%%%%%%%%%%%
\label{sec:conclusion}

In this work we have presented a general mathematical formulation of the cloud offloading problem, tailored to robotic systems. Our formulation as a Markov Decision Problem is both general and powerful. We have demonstrated deep reinforcement learning may be used within this framework effectively, outperforming common heuristics. However, we wish to emphasize that RL is likely an effective choice to optimize offloading policies even for modifications of the offloading problem as stated. 

\paragraph*{Future Work}

While there are many theoretical and practical avenues of future work within the cloud robotics setting (and more specifically within offloading), we wish to herein emphasize two problems that we believe are important for improved performance and adoption of cloud robotics techniques. First, we have characterized the offloading problem as an MDP, in which factors such as latency correspond to costs. However, for safety critical applications such as self-driving cars, one may want to include hard constraints, such as a bounding the distribution of latency times. This approach would fit within the scope of Constrained MDPs \cite{altman1999constrained}, which has seen recent research activity within deep reinforcement learning \cite{chow2018lyapunov,achiam2017constrained}.

Secondly, we have dealt with input streams that are independent of our decisions in this work. However, the input streams that robotic systems receive are a consequence of the actions that they take. Therefore, a promising extension to improve performance of cloud robotics systems is considering the offloading problem and network characteristics during action selection (e.g., planning or control). Conceptually, this is related to active perception \cite{bajcsy1988active}, but also incorporates information about network conditions or input stream stale-ness. 

% \paragraph*{Note to Practitioners}

%% Use plainnat to work nicely with natbib.

{
% \renewcommand{\baselinestretch}{.9}

% \clearpage
\bibliographystyle{plainnat}
\bibliography{main}
}

\end{document}

%% file: sections/preamble.tex
\usepackage{dsfont}
\usepackage{microtype}
\usepackage{graphicx}
\usepackage{subfig}
\usepackage{booktabs} % for professional tables
\usepackage{xspace}
\usepackage[dvipsnames]{xcolor}
\usepackage{multirow}
\usepackage{hhline}
\usepackage{nicefrac}

\usepackage{footnote}
\makesavenoteenv{tabular}
\makesavenoteenv{table}

\usepackage{epsfig}
\usepackage{amssymb}
\usepackage{amsmath}
\usepackage{amsfonts}

\usepackage{stackengine}
\usepackage{algorithmic,comment}
\usepackage[ruled,vlined,commentsnumbered,titlenotnumbered]{algorithm2e}
\usepackage{float}

\usepackage{colortbl}
\usepackage{xcolor}
\definecolor{LightCyan}{rgb}{0.88,1,1}

\usepackage{color}

\newcommand{\jetson}{Jetson Tx2}
\newcommand{\nvidia}{NVIDIA}

\newcommand{\expec}{\mathbb{E}}

\newcommand{\yhat}{\hat{y}}
\newcommand{\loss}{\mathcal{L}(y^t, \hat{y}^t)}

\newcommand{\fcloud}{f_\mathrm{cloud}}

\newcommand{\yconfcloud}{\mathtt{conf}_{\mathrm{cloud}}}

\newcommand{\yconf}{\mathtt{conf}}

\newcommand{\taucloud}{\tau_{\mathrm{cloud}}}

\newcommand{\trobot}{t_{\mathrm{robot}}}
\newcommand{\tcloud}{t_{\mathrm{cloud}}}

\newcommand{\yhatcloud}{\hat{y}_{\mathrm{cloud}}}

\newcommand{\budget}{N_{\mathrm{budget}}}
\newcommand{\Nbudgetfrac}{\frac{N_{\mathrm{budget}}}{T}}

\newcommand{\MDPoffload}{\mathcal{M}_{\mathrm{offload}}}
\newcommand{\pioffload}{\pi_{\mathrm{offload}}}
\newcommand{\Soffload}{\mathcal{S}_{\mathrm{offload}}}
\newcommand{\Aoffload}{\mathcal{A}_{\mathrm{offload}}}
\newcommand{\Poffload}{\mathcal{P}_{\mathrm{offload}}}
\newcommand{\Roffload}{R_{\mathrm{offload}}}
\newcommand{\aoffload}{a_{\mathrm{offload}}}
\newcommand{\soffload}{s_{\mathrm{offload}}}
\newcommand{\cost}{\mathtt{cost}}

\newcommand{\alphaaccuracy}{\alpha_{\mathrm{accuracy}}}
\newcommand{\betacost}{\beta_{\mathrm{cost}}}

\newcommand{\pioffloadrandom}{\pi_{\mathrm{offload}}^{\mathrm{random}}}

\newcommand{\pioffloadrobot}{\pi_{\mathrm{offload}}^{\mathrm{all-robot}}}
\newcommand{\pioffloadcloud}{\pi_{\mathrm{offload}}^{\mathrm{all-cloud}}}
\newcommand{\pioffloadheuristic}{\pi_{\mathrm{offload}}^{\mathrm{robot-heuristic}}}
\newcommand{\pioffloadRL}{\pi_{\mathrm{offload}}^{\mathrm{RL}}}

\newcommand{\argmax}{\mathop{\rm argmax}}

\newcommand{\maskrcnn}{{Mask R-CNN~}}

\newtheorem{problem}{Problem}

\newcommand{\frobot}{f_\mathrm{robot}}

\newcommand{\yconfrobot}{\mathtt{conf}_{\mathrm{robot}}}

\newcommand{\taurobot}{\tau_{\mathrm{robot}}}

\newcommand{\yhatrobot}{\hat{y}_{\mathrm{robot}}}

%% file: sections/evaluation.tex
% The principal goal of our evaluation is to benchmark
% the performance of the our proposed RL cloud offloading policy in simulations
% and hardware experiments with real computer vision DNNs. Our evaluation
% shows:

We benchmark our proposed RL-based cloud offloading policy within a realistic and representative setting for cloud robotics. Specifically, we focus on a face detection scenario using cutting edge vision DNNs. This scenario is prevalent in robotics applications ranging from search and rescue to robots that assist humans in commercial or industrial settings. More generally, it is representative of an object detection task that is a cornerpiece of virtually any robotics perception pipeline. We test this system with both a simulated input image stream with controlled temporal coherence as well as on a robotic hardware platform with real video streams, and find that in both cases, the RL policy intelligently, and sparingly, queries the cloud to achieve high prediction accuracy while incurring low query costs, outperforming baselines.

% \begin{enumerate}
%     \item RL intelligently, but sparingly, learns to query the cloud
%           to accrue high reward, low model query cost, and high prediction accuracy
%           (Figs. \ref{fig:sim_reward} - \ref{fig:sim_cost}).

%       \item \mpmargin{Similar performance on live video streams captured from an embedded compute platform optimized for AI applications}{ill-formed sentence, unclear} (Section \ref{subsec:hardware_exp}).
% \end{enumerate}

%We rigorously evaluated the performance of an RL offloader
%in both simulation, using real computer vision DNNs, and with
%hardware experiments using 

%\textit{Simulation scenario:}

\paragraph*{Face-detection Scenario}
We formulate this scenario, depicted in Fig. \ref{fig:facenet_demo}, in terms of the general abstractions we introduced in Section \ref{sec:prob}. Here, the sensory input stream is a video, where each $x^t$ is a still frame from that video. To avoid training a policy over the large image space directly, we choose the feature encoding $\phi$ that is used in the state space of the offloading MDP to be the sum of absolute differences between sequential frames. 
For the on-robot prediction model $\frobot$, we use a combination of FaceNet \cite{schroff2015facenet}, a widely-used pre-trained face detection model which embeds faces into embedding vectors, together with an SVM classifier over these embeddings. This model has seen success for face detection in live streaming video on embedded devices \cite{openface}. For the cloud model $\fcloud$, we use a human oracle, which always gives an accurate prediction with a high confidence measure. We used a zero-one loss function to measure the accuracy of the predictions, with $\loss = 1$ if the prediction was incorrect, and $\loss = 0$ if it was correct.

    We choose the reward function to balance prediction accuracy and minimize onboard computation, as well as queries to the human operator through the network. The cost of past robot model and cloud queries, denoted by actions $0,1$, was set to zero ($\cost(0) = \cost(1) = 0$), while the robot model cost was set to $\cost(2) = 0.4$ and the cost of the cloud model was chosen to be $\cost(3) = 8.0$, to especially penalize querying the human oracle who will have limited bandwidth. We tested with different weightings in the reward function (Eqn. \ref{eq:twostage_reward}), and found $\alphaaccuracy = 1.0$ and $\betacost = 7.0$ to yield performance for our specific cost setup, and therefore report results for this parameter setting. These costs were chosen to incentivize reasonably rational behavior; in real robotic systems they could be computed through an economic cost-benefit analysis\footnote{We provide the offloading simulation environment, robot and cloud FaceNet models, and MDP dynamics outlined in Eqns. \ref{eq:twostage_state} - \ref{eq:twostage_reward}, as a standard OpenAI \textit{gym} \cite{brockman2016openai} environment at \url{https://github.com/StanfordASL/cloud_robotics}.}.

%[removed for anonymous peer review].}

\paragraph*{Offloading Policy Architecture}
% Policy architecture
% We represent the RL offloading policy as a deep neural network and train it using the Advantage Actor-Critic (A2C) algorithm \cite{mnih2016asynchronous}.
In practice, the input query sequence may show time-variant patterns and the MDP may become nonstationary if the agent only knows the current state. To address this problem using a recurrent policy, we use a Long Short Term Memory (LSTM) \cite{hochreiter1997long} as the first hidden layer in the offloader policy
to extract a representation over a short history of states. In particular, the actor (or critic) DNN has a LSTM first layer of 64 units, a fully-connected second layer of 256 units, and a softmax (or linear) output layer. 
%Also, we train actor and critic DNNs separately, instead of sharing lower layers among them. 
We softly enforce the action constraint of disallowing the offloading action when the budget has depleted by having action 3 map to action 2 when $\budget = 0$.

We use standard hyper-parameters for A2C training, with an orthogonal initializer and RMSprop gradient optimizer.
Specifically, we set actor learning rate to $10^{-4}$, critic learning rate to $5e-5$, minibatch size $20$, entropy loss coefficient $0.01$, and gradient norm clipping $40$. 
We train A2C over 1 million episodes, with discount factor $0.99$ and episode length $T=80$. We observed stable convergence after $350,000$ episodes, consistent over different weightings of the accuracy and loss terms in the reward.

A key aspect of this problem is how the \textit{coherence} of the input stream allows the offloading policy to leverage cached predictions to avoid excessively querying the cloud model. In order to test this, we applied the deep RL approach in two scenarios: a synthetic stream of images where coherence was controlled, as well as an on-hardware demo which used real video data. In the following subsections, we detail the training and testing procedure for each scenario, and discuss the results.

\subsection{Synthetic Input Stream Experiments}
\label{subsec:rl_offloader}
To model the coherence of video frames we observed experimentally, we divided an episode of $T$ steps
into ``coherent'' sub-intervals, where only various frames of the \textit{same person} appear within one contiguous sub-interval, albeit with different background and lighting conditions. 
Then, the input stochastically switches to a new individual, who could be unknown to the robot model. As such, we simulate a coherent stream of faces which are a diverse mixture of known and unknown faces to the robot, as shown at the top of Fig. \ref{fig:facenet_demo}.
The length of a coherence interval was varied between $\nicefrac{1}{10}-\nicefrac{1}{12}$ of an episode duration $T$ to show a diversity of faces in an episode.

Each training trace (episode of the MDP) lasted $T = 80$ steps where a face image (query $x^t$) arrived at each timestep $t$. To test RL on a diverse set of network usage limits, we randomly sampled a query budget $\Nbudgetfrac \in [0.10,0.20,0.50,0.70,1.0]$ at the start of each trace.

\begin{figure}[t]
    \centering
	\subfloat{
    	\includegraphics[width=0.5\textwidth]{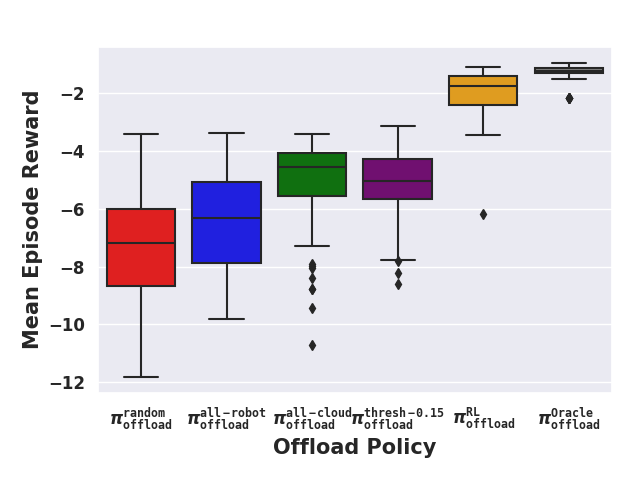}
    }
    \caption{RL beats benchmark offloading policies by over $2.6\times$ in diverse test episodes over a mixture of network conditions.}
    \label{fig:sim_reward}
\end{figure}

\begin{figure}[t]
    \centering
	\subfloat{
        \includegraphics[width=0.5\textwidth]{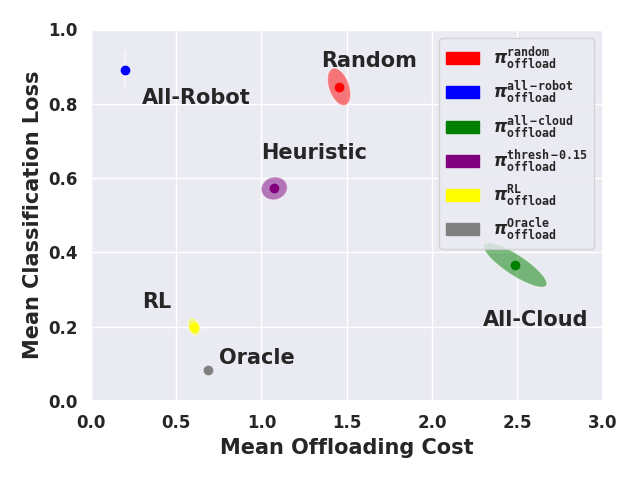}
    }
    \caption{The reward trades off offloading costs, which penalize network and cloud usage, with classification loss.}
    \label{fig:sim_cost}
\end{figure}

\paragraph*{Evaluation}
We evaluated the RL policy and the benchmarks on 100 diverse testing traces each,
where the face pictures present in each trace were distinct from those in the set of training
traces. To test an offloader's ability to adapt to various network bandwidth constraints, 
we evaluated each trace with four trials on each query budget fraction in $\Nbudgetfrac \in [0.10,0.20,0.50,0.70,1.0]$, simulating budgets in highly-constrained to unconstrained networks. 

We show RL test results for the same representative reward function parameters described above in Section \ref{subsec:rl_offloader}.

%Our RL offloader has significantly better reward than benchmark policies, and learns to sparingly query the costly cloud model only when locally uncertain, otherwise using the cheaper local robot model.
%Our RL offloader

\paragraph*{RL Intelligently, but Sparingly, Queries the Cloud}
Figure \ref{fig:sim_reward} shows the distribution of rewards attained by
the different offloader policies on all test traces, 
where our RL approach is depicted in the yellow boxplot.
Then, we break down the mean episode reward into its components of prediction accuracy and offloading cost, and show the mean performance over all test traces for each policy in Fig. \ref{fig:sim_cost}.

Benchmark policies of random-sampling ($\pioffloadrandom$), all-robot compute ($\pioffloadrobot$), periodic cloud-compute ($\pioffloadcloud$), and the \textit{best} confidence-threshold based heuristic policy ($\pioffloadheuristic$) are shown in the left four boxplots (red to purple). An oracle upper-bound solution, which is unachievable in practice since it \textit{perfectly} knows the robot and cloud predictions and future timeseries $x^t$, is depicted in gray in Figs. \ref{fig:sim_reward} - \ref{fig:sim_cost}.

Fig. \ref{fig:sim_reward} shows that RL has at least $2.6\times$ higher median episode reward than the benchmarks, and is competitive with the upper-bound oracle solution, achieving $0.70\times$ its reward.
This is because the RL policy sparingly queries the costly cloud model, in contrast to an all-cloud policy that incurs significantly higher model query and network cost, as shown in Fig. \ref{fig:sim_cost}, which plots the mean reward terms and the 95\% standard error estimates to display our certainty in the mean estimate.
Essentially, RL learns to judiciously query the cloud when the robot model is highly uncertain, 
which allows it to improve the overall system accuracy and achieve a low prediction loss (Fig. \ref{fig:sim_cost}). Interestingly, it has better prediction accuracy than an ``all-cloud'' scheme since bandwidth limits cause this policy to periodically, but sparsely, sample the cloud and hold past cloud predictions. RL learns to conserve precious cloud queries when the query budget $\Nbudgetfrac$ is low and use them when they help prediction accuracy the most, thereby achieving low prediction error in Fig. \ref{fig:sim_cost}. 

% \mpmargin{Note}{I still find this paragrpah weird: you seem to tell the reader that we have computed a pareto forntier, and then you say we ere to lazy to do it} that the oracle is the optimal policy for our particular weighting of the two cost terms. As the relative weighting of these two terms is varied, it traces the Pareto frontier. Moreover, note that the all-robot policy is one extremum of the Pareto frontier, as the relative cost of offloading relative to the classification loss goes to infinity. We have only visualized one Oracle policy (as opposed to the entire Pareto frontier) due to the high associated computational cost.

%\begin{figure}[h]
%   \centering
%	\subfloat{
%        \includegraphics[width=0.5\textwidth]{pics/plots/RL_2.pdf}
%    }
%    \caption{Prediction accuracy of various offloading policies.}
%    \label{fig:sim_loss}
%\end{figure}

% FACENET PICTURE
%%%%%%%%%%%%%%%%%%%%%%%%%%%%%%%%%%%%

\subsection{Hardware Experiments}
\label{subsec:hardware_exp}
%noindent \textbf{Hardware Experiments:}
Inspired by deep RL's promising performance on synthetic input streams, we built an RL offloader that runs on 
the NVIDIA \textit{\jetson} embedded computer, which is optimized for deep learning 
and used in mobile robotics.
The RL offloader takes in live video from the Jetson camera and runs the small \texttt{nn4.small2.v1}\footnote{available at \url{https://cmusatyalab.github.io/openface/models-and-accuracies/}}
FaceNet DNN from the popular  OpenFace project \cite{openface} and an SVM classifier on selected frames as the robot model. OpenFace \cite{openface} provides four pre-trained FaceNet models, and we chose to use the smallest, fastest model on the robot, to consider cases where a robot has limited hardware.
The smallest model has half the parameters and is $1.6\times$ faster on a GPU than the largest FaceNet model \cite{openface}.

If the RL agent deems a face needs to be offloaded due to local uncertainty, it
can either send the concise Facenet embedding (128-dimensional byte vector) or the face image to a central server
that can run a larger FaceNet DNN and/or SVM classifier trained on many more humans of interest as the cloud model.
%Alternatively, it can offload selected video frames (images) to be classified by a human, and the returned result is used as the robot's prediction.
We use OpenCV for video frame processing, a PyTorch \texttt{nn4.small2.v1} OpenFace model, and TensorFlow \cite{abadi2016tensorflow} for the RL offloader neural network policy to achieve real-time processing performance on the embedded Jetson platform.

%\subsubsection{Data Collection}
\paragraph*{Data Collection}

We captured 6 training and 3 testing videos 
spanning 9 volunteers, consisting of over 2600 frames of video
where offloading decisions could be made.
The nine volunteers were known to the robot model but our dataset had 18 distinct, annotated
people. Collectively, the training and test datasets showed diverse scenarios, ranging
from a single person moving slowly, to dynamic
scenarios where several humans move amidst background action.

\begin{figure}[t]
    \centering
    \subfloat[FaceNet on a live video stream.]{
        \includegraphics[width=0.45\linewidth]{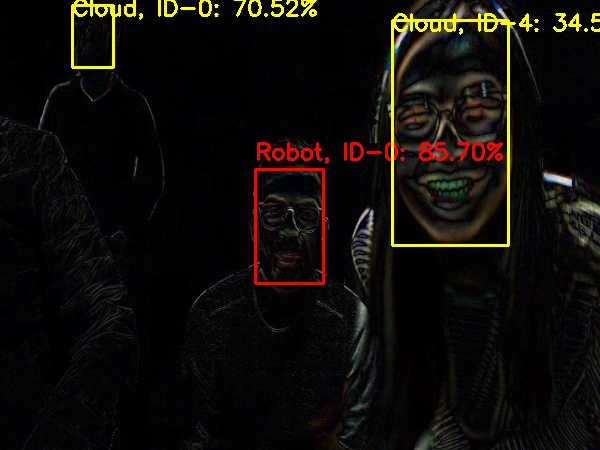}
    }
    \subfloat[Offload yellow faces.]{
        \includegraphics[width=0.45\linewidth]{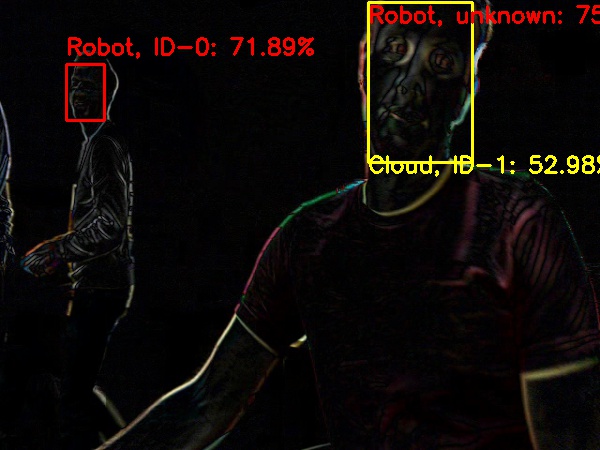}
    }
    \caption{\textbf{Hardware Experiments:} Our offloader, depicted on frame pixel differences, interleaves FaceNet predictions on a robot (red box) and cloud model (yellow) when uncertain.}
    \label{fig:video_example}
\end{figure}

The Jetson was deployed with a robot FaceNet model and an SVM trained on a subset of images
from only $9$ people, while the cloud model was trained on several more images of all $9$ volunteers
to be more accurate. The state variables in Eq. \ref{eq:twostage_state} were input to our Deep RL offloader, where the sum of frame \textit{differences } $\sum_{\mathrm{pixels}} | x^t - x^{t-1} |$,  rather than a full image, was an important indicator of how quickly video content was changing. Frame differences are depicted in Fig. \ref{fig:video_example}, which helps the offloader subtract background noise and hone in on rapidly-changing faces.

% data collection

% use of difference frames

% how many known people

% availability of videos, with annotated results publicly available at

%\subsubsection{Experimental Insights}
\paragraph*{Evaluation and Discussion}
As expected, the trained RL offloader queries the cloud
when robot model confidence is low, the video is chaotic (indicated by 
large pixel difference scores), and several hard-to-detect faces appear in the background.

However, it weights such factors together to decide an effective policy and hence
is significantly better than a confidence-threshold based heuristic. In fact, the 
decision to offload to the cloud is only 51.7\% correlated (Spearman correlation coefficient) with robot model confidence, 
showing several factors influence the policy.

In our hardware experiments on real streaming video, our offloader achieved
$1.6\times$ higher reward than an all-robot policy, $1.35\times$ better reward than an all-cloud policy, and $0.82\times$ that of an oracle upper bound solution. Further, it attained $1.1-1.5\times$ higher reward
than confidence-based threshold heuristics, where we did a linear sweep over threshold confidences, which depend on the specific faces in the video datasets.

Our video dataset, robot/cloud models, and offloader code will be made publicly  available. In particular, videos of the offloading policy show how robot computation can be effectively juxtaposed with cloud computation in Fig. \ref{fig:video_example}. Finally, since the offloader has a concise state space and does not take in full images, but rather a sum of pixel differences as input, it is extremely small ($900$ KB). Essentially, it is an order of magnitude smaller than even optimized vision DNNs (Table \ref{fig:DNN_table}), allowing it to be scalably run on a robot without interfering in perception or control.

% insights: correlation of SVM confidence with offloading decision

% RL offloads when several detections in a scene, frame changes a lot, robot model confidence poor

% beats a threshold-based policy